\documentclass[
	a4paper, 
	10pt, 
	unnumberedsections, 
	twoside, 
]{LTJournalArticle}

\usepackage{amsfonts}

\runninghead{} 

\footertext{\textit{Creativity in AI as Emergence from Domain-Limited Generative Models}} 

\title{Creativity in AI as Emergence from Domain-Limited Generative Models} 

\author{%
	\textbf{Corina Chutaux}
    }

\date{Sorbonne University\\ corina@chutaux.com}

\renewcommand{\maketitlehookd}{%
	\begin{abstract}
		\noindent Creativity in artificial intelligence is most often addressed through evaluative frameworks that aim to measure novelty, diversity, or usefulness in generated outputs. While such approaches have provided valuable insights into the behavior of modern generative models, they largely treat creativity as a property to be assessed rather than as a phenomenon to be explicitly modeled. In parallel, recent advances in large-scale generative systems, particularly multimodal architectures, have demonstrated increasingly sophisticated forms of pattern recombination, raising questions about the nature and limits of machine creativity.
        This paper proposes a generative perspective on creativity in AI, framing it as an emergent property of domain-limited generative models embedded within bounded informational environments. Rather than introducing new evaluative criteria, we focus on the structural and contextual conditions under which creative behaviors arise. We introduce a conceptual decomposition of creativity into four interacting components-pattern-based generation, induced world models, contextual grounding, and arbitrarity, and examine how these components manifest in multimodal generative systems. By grounding creativity in the interaction between generative dynamics and domain-specific representations, this work aims to provide a technical framework for studying creativity as an emergent phenomenon in AI systems, rather than as a post hoc evaluative label.

	\end{abstract}
}

\begin{document}

\maketitle 


\section{Introduction}

Recent progress in generative artificial intelligence has significantly expanded the range and quality of machine-generated artifacts. Advances in transformer-based language models, diffusion-based image synthesis, and multimodal architectures have enabled systems capable of producing coherent text, detailed images, and cross-modal associations at unprecedented scales. These developments have naturally revived longstanding questions about creativity in artificial intelligence: whether such systems can be considered creative, how creativity should be defined in computational terms, and how it might be evaluated or engineered.

Most contemporary discussions of creativity in AI adopt an evaluative stance. Creativity is typically defined through criteria such as novelty, diversity, usefulness, or surprise, and models are assessed by measuring how well their outputs satisfy these properties \cite{r3}, \cite{r23}. While this perspective has proven effective for benchmarking and comparison, it leaves open a more fundamental question: how creativity emerges from the internal dynamics of generative systems themselves. In practice, creativity is often inferred from outputs after generation, rather than treated as a phenomenon to be modeled within the generative process.

At the same time, modern generative models already demonstrate behaviors that go beyond simple pattern replication. Large language models exhibit flexible recombination of concepts, contextual adaptation, and the ability to sustain coherent discourse across extended interactions \cite{r20}. Multimodal systems integrate textual and visual representations, enabling cross-domain associations that resemble aspects of human creative practice. These behaviors suggest that creativity in AI may arise not from explicit creative rules, but from the interaction between learned representations, generative dynamics, and the informational environments in which models are trained.

Earlier attempts to emulate artistic creativity in artificial systems often framed creativity as a goal-oriented decision process, focusing on saliency, emphasis, or subjective choice within a predefined scene. For example, approaches such as \cite{r5} model portrait painting by identifying which elements of a scene should be emphasized, relying on hierarchical saliency structures and heuristic decision rules. While such models provide insights into perceptual attention and compositional preferences, they operate primarily at the level of output selection and do not address the emergence of novel forms from internal representational dynamics. As a result, creativity is treated as a sequence of decisions applied to known elements, rather than as a generative process arising from the interaction between structured constraints, learned representations, and contingent variation.

This paper adopts the position that creativity in AI should be studied as an emergent property of generative systems embedded within bounded domains of knowledge. Rather than asking whether a model is creative in absolute terms, we investigate how creativity arises relative to a specific informational world defined by training data, representational structure, and modality. In this view, training corpora are not merely data resources but define coherent epistemic environments that shape what a model can represent, recombine, and invent.

Building on this perspective, we propose a conceptual framework that decomposes creativity into four interacting components. First, pattern-based generation captures the ability of models to learn and recombine structural regularities present in the data. Second, induced world models refer to the internal representations that organize concepts, relations, and semantic structures within the domain. Third, contextual grounding captures the influence of cultural, historical, or domain-specific regularities that give meaning to generated artifacts. Finally, arbitrarity accounts for the flexibility and non-determinism that allow models to depart from dominant patterns and explore less probable associations.

This framework is particularly relevant for multimodal generative systems, which combine multiple representational spaces—such as text and images—into a shared generative process. While multimodality does not provide experience in the human sense, it substantially enriches the space of associations available to the model, enabling more complex forms of contextual and cross-domain generation. We argue that such systems provide a natural setting for studying creativity as a computational phenomenon emerging from the interaction of structure, context, and generative freedom.

The contributions of this paper are threefold. First, we provide a technical perspective on creativity in AI that shifts the focus from evaluation to emergence within generative systems. Second, we introduce a domain-limited framework for analyzing creativity as a property relative to bounded informational environments rather than as a universal capability. Third, we outline a structural decomposition of creativity that can be connected to concrete generative mechanisms in modern AI models. Together, these contributions aim to support a more precise and generative understanding of creativity in artificial intelligence.


\section{State of the Art}

\subsection{Creativity in AI: From Definitions to Operational Criteria}

Research on computational creativity has historically focused on defining creativity through a set of evaluative criteria, most commonly novelty, value, usefulness, and surprise \cite{r6}, \cite{r1}. These criteria have provided a shared conceptual basis for comparing heterogeneous generative systems and have enabled the development of evaluation protocols based on human judgments, psychometric tests, or automated proxy metrics \cite{r17}. Despite the evolution of generative architectures, these foundational criteria continue to structure much of the contemporary discourse on creativity in artificial intelligence.
However, while these definitions remain influential, they primarily address creativity as an observable property of outputs rather than as a computational process to be explicitly modeled. As a result, many recent contributions revisit established creativity definitions without proposing fundamentally new generative mechanisms or formal models that aim to emulate creativity at the algorithmic level.

\subsection{Creativity Assessment in Modern Generative Models}

With the advent of large-scale generative models, particularly large language models and diffusion-based image generators, recent research has increasingly concentrated on the evaluation of creative behavior. Typical approaches include divergent-thinking benchmarks, diversity and originality metrics, human preference studies, and learned evaluators trained to predict creativity-related judgments \cite{r20}.
While these evaluation-driven approaches provide valuable insights into the apparent creative capacities of modern models, they remain largely post hoc. Creativity is inferred after generation, based on external criteria, rather than embedded as an explicit objective or structural property of the generative process itself. Consequently, creativity is treated as something to be detected or validated, not as something to be computationally instantiated.

\subsection{Generative Architectures and Explicit Creativity Objectives}

Only a limited subset of work has attempted to incorporate creativity directly into the generative objective. Among these, Creative Adversarial Networks \cite{r9} represent a notable example, as they explicitly introduce a learning signal encouraging deviation from known stylistic distributions rather than faithful imitation. Related ideas also appear in novelty search, intrinsic motivation, and quality-diversity algorithms, where exploration and deviation are rewarded as first-class objectives.
Despite these contributions, such approaches remain relatively isolated within the broader landscape of generative modeling. Most modern architectures prioritize likelihood maximization, reconstruction fidelity, or task performance, implicitly relying on scale and data diversity to produce creative-looking outputs rather than modeling creativity as an explicit computational phenomenon.

\subsection{Pattern-Based Creativity and the Limits of Current Models}

Contemporary generative models excel at learning and recombining patterns across large datasets. In text, this manifests as fluent syntactic and semantic interpolation; in images, as stylistic blending and structural recomposition. This form of creativity—here referred to as pattern-based creativity—is a direct consequence of representation learning in high-dimensional latent spaces.
However, this focus on patterns raises unresolved questions about the nature of creativity being modeled. Pattern recombination alone does not account for broader contextual, cultural, or epistemic structures that shape creative production. In particular, most models lack an explicit notion of the world in which creativity takes place: the historical, cultural, or conceptual environment that constrains and gives meaning to creative acts.

\subsection{Domain-Limited Generative Models and Situated Creativity}

An underexplored dimension in the literature concerns creativity as an emergent property of systems embedded within a bounded informational domain. Rather than treating training data as a neutral resource, a domain-limited perspective considers the dataset as defining a coherent epistemic environment that shapes the model’s internal representations, generative dynamics, and creative possibilities.
From this viewpoint, creativity is not evaluated in absolute terms but relative to the conceptual space induced by the domain. A model trained on a historically or culturally bounded corpus is expected to exhibit creativity that emerges from that specific world of representations, norms, and constraints. This shift reframes creativity as a property of situated generative systems, rather than as a universal capability measured independently of context.

\subsection{Positioning of the Present Work}

In summary, recent work on creativity in AI predominantly emphasizes definitional refinement and evaluative methodologies, while comparatively few approaches address creativity as a computational process to be explicitly modeled within generative systems. Moreover, existing frameworks often abstract away the role of domain-specific world structures, treating creativity as a task-agnostic property.
The present work addresses this gap by investigating creativity in AI as an emergent phenomenon arising from domain-limited generative models. Rather than proposing new evaluative criteria, it focuses on the interaction between learned patterns, internal world models, and contextual structure, aiming to formalize creativity as a property of generative systems embedded within bounded informational environments.

\section{Conceptual and Mathematical Framework}

\subsection{Empirical Decomposition of Creativity’s Origin}

The framework proposed in this work did not originate from an abstract theoretical model of creativity, nor from a neuroscientific account of artistic cognition. Instead, it emerged from an empirical and inductive analysis of creative practices, informed by discussions with artists, examination of art-historical cases, and comparative observation across creative domains. These observations consistently suggest that artistic creativity cannot be reduced to a set of internal cognitive rules or localized mechanisms. Rather, creative production appears as a situated phenomenon, shaped by contextual conditions, individual worldviews, learned structures, and contingent events.
Historical examples illustrate the constitutive role of context in creative production, where political, social, and cultural environments directly influence artistic expression. At the same time, artists systematically externalize their individual systems of beliefs, values, and interpretations of the world through their chosen media, resulting in coherent and recognizable worldviews. Creative processes also repeatedly involve moments of contingency—errors, accidental discoveries, or unplanned deviations—that retrospectively acquire meaning and become integral to artistic or scientific practice. Finally, all creative activity relies on the acquisition, internalization, and transformation of learned patterns accumulated through sustained exposure and practice.
These recurring empirical observations motivate the decomposition of creativity introduced in this paper, consisting of four interacting components: contextual grounding (Zeitgeist), induced world models (Weltanschauung), pattern-based generation (Patternism), and Arbitrarity.

\subsection{Conceptual Definitions}

This section introduces the four conceptual components used to characterize creativity in domain-limited generative systems. These components are not intended as psychological faculties or evaluative criteria, but as structural dimensions describing how creative behavior emerges from generative processes embedded in bounded informational environments.\\

\textbf{Zeitgeist (Contextual Grounding)}\\

Zeitgeist refers to the contextual regularities induced by the cultural, historical, and informational environment in which a generative system is trained. In computational terms, it corresponds to the statistical, semantic, and stylistic distributions that dominate a given domain. These distributions define what is typical, expected, or coherent within that environment. Creativity relative to the zeitgeist does not require departure from the domain itself, but rather operates within its implicit norms while allowing for non-trivial variation.\\

\textbf{Weltanschauung (Induced World Model)}\\

Weltanschauung denotes the internal organization of meaning learned by a generative model from its training data. It captures how concepts, relations, and representations are structured into a coherent internal model of the domain. Unlike zeitgeist, which reflects external regularities, Weltanschauung refers to the model’s induced representational geometry: its implicit ontology, conceptual associations, and semantic coherence. Creative generation depends on this internal structure, as it constrains how elements can be recombined in meaningful ways.\\

\textbf{Patternism (Learned Structural Regularities)}\\

The notion of patternism adopted in this work is consistent with computational theories of intelligence that characterize cognition as the detection, abstraction, and transformation of patterns across representational levels (e.g., Goertzel,\cite{r25}).Therefore, patternism describes the acquisition and manipulation of recurring structures present in the data, including syntactic, stylistic, visual, or compositional patterns. In generative models, patternism arises naturally from representation learning and optimization over large datasets. Creative behavior at this level corresponds to the recombination, transformation, or distortion of learned patterns, producing outputs that are novel relative to specific instances while remaining consistent with domain-level regularities.
Patternism (Learned Structural Regularities) describes the acquisition and manipulation of recurring structures present in the data, including syntactic, stylistic, visual, or compositional patterns. In generative models, patternism arises naturally from representation learning and optimization over large datasets. Creative behavior at this level corresponds to the recombination, transformation, or distortion of learned patterns, producing outputs that are novel relative to specific instances while remaining consistent with domain-level regularities.\\

\textbf{Arbitrarity (Contingent Deviation)}\\

Arbitrarity captures the role of contingent deviation in creative processes. It refers to the emergence of variations that are not strictly determined by dominant patterns or internal structures, but arise through unplanned or low-probability events. In computational systems, arbitrarity is typically instantiated through stochastic mechanisms, but its creative function lies not in randomness itself, rather in enabling exploration beyond high-likelihood modes of generation. Arbitrarity allows generative systems to produce outcomes that can later be integrated meaningfully within the domain.

\section{Mathematical Formalization of Creativity}

This section formalizes the proposed decomposition of creativity into (i) \textit{Weltanschauung} (individual worldview), (ii)\textit{Patternism} (individual pattern-learning and recombination), (iii) \textit{Zeitgeist} (historical-cultural signal at the scale of a century), plus (iv) an \textit{arbitrarity} term modeled as a random constant (a stochastic shock) rather than a component with its own internal structure.

\subsection{Notation and indexing}

We use three indexing axes:

\begin{itemize}
	\item	\textbf{Individual index}: \textit{I} (an individual / agent / model instance in a given run)
	\item \textbf{Century (macro-historical) index}: \textit{S} (e.g., the 18th century)
	\item \textbf{Time indices}:
               \subitem\textit{t}: an individual-time index (e.g., “life time” or iteration-time for a system)
               \subitem\textit{i}: a discrete year-like index used for aggregation inside a century

\end{itemize}

We denote the overall creativity (as an emergent generative capacity within a domain) by:

\begin{center}
    \[ C_{I,S}(t) \]
\end{center}

i.e., creativity produced by an “individual” \textit{I} embedded in (or trained within) a historical domain \textit{S}, at time \textit{t}.

\subsection{Patternism as an individual-time aggregation}

\textbf{Patternism accumulates over an individual horizon} (a lifespan, or algorithmically an exposure/training horizon).\\

We formalize it as:

\begin{equation}
P_I(t) = \sum_{k=0}^{t} p_I(k)
\end{equation}

where $\textit{p}_I(\textit{k})$ is the incremental contribution of pattern acquisition / recombination at step \textit{k}. In “human” reading, \textit{t} can be the age (years lived); in the algorithmic reading, \textit{t} can be training steps / epochs / iterations (or any monotonically increasing exposure variable).
\\So $\textit{p}_I(\textit{t})$ grows as the system internalizes regularities (composition, style priors, motifs) and becomes able to recombine them.

\subsection{Zeitgeist as a century-scale aggregation}

Zeitgeist expresses a \textit{century-wide accumulation} (100 years) which can be formulized as following: 

\begin{equation}
Z_S = \sum_{i=0}^{99} z_S(i)
\end{equation}

where $z_S(i)$ is the “year-level” contribution to the historical-cultural signal during year \textit{i} of century \textit{S}. In practice (for a corpus-based model), $\textit{Z}_S$ is not measured by calendar years per se; it is an aggregate of the domain’s distributions (textual topics, aesthetic conventions, iconography, lexicon, composition rules) associated with that century.
So $\textit{Z}_S$  is a domain operator: it characterizes the historical manifold the model is trained on.

\subsection{Weltanschauung as a structured individual function}

The worldview is a structured sum of subcomponents which can be represented as: 

\begin{equation}
W_I(t) = B_I(t) + V_I(t) + E_I(t)
\end{equation}

\begin{itemize}
	\item $\textit{B}_\textit{I}(\textit{t})$: belief-structure (internalized propositions, causal expectations, narratives)
	\item $\textit{V}_\textit{I}(\textit{t})$: value-structure (preferences, salience weights, normative attractors)
	\item $\textit{E}_\textit{I}(\textit{t})$: experience-structure (accumulated exposures, perceptual and cultural traces)
\end{itemize}

\subsection{Arbitrarity as an uncontrollable constant (shock term)}

\textbf{Arbitrarity is not a dimension with its own equation}. It is a constant term, a rare/irregular perturbation that cannot be planned, but can redirect trajectories (accidents, mistakes, lucky discoveries, serendipity, noise exploited as signal).

We model it as:

\begin{equation}
\varepsilon \sim \mathcal{D}
\end{equation}

where \textit{D} is some distribution or more simply an additive constant for the purpose of the decomposition:

\begin{equation}
\varepsilon \in \mathbb{R} \quad (uncontrolled)
\end{equation}

Arbitrarity \textbf{is not the algorithmic randomness alone}; it is the more general phenomenon that “unexpected deviations” can become creatively meaningful.

\subsection{The combined creativity equation}

To account for unequal contributions of the different components, the formulation can be extended by introducing scalar weights:

\begin{equation}
C_{I,S}(t) = \alpha\, W_I(t) + \beta\, P_I(t) + \gamma\, Z_S + \varepsilon
\end{equation}

where $\alpha, \beta, \gamma \in \mathbb{R}^+$ modulate the relative influence of individual worldview, accumulated patterns, and historical context, respectively. This weighted formulation does not define a creativity metric in an evaluative sense. Rather, it specifies a generative decomposition describing the structural conditions under which creative behavior may emerge within a domain-limited system. The introduction of weights allows the framework to accommodate different generative regimes without altering its conceptual core.

\subsection{Scope of the Formalization}

This formalization is intentionally generative rather than evaluative. It does not propose a creativity score, benchmark, or optimization target. Instead, it provides a \textbf{mathematical scaffold} for constructing and analyzing domain-limited generative systems in which creative behavior may emerge under historically and individually grounded conditions.

\section{Model and Methodology}

This section describes a concrete instantiation of the mathematical framework introduced in Section 3 through a series of generative experiments. As an initial baseline, the visual corpus was first trained using a Deep Convolutional Generative Adversarial Network (DCGAN) operating in a purely unimodal setting. In this configuration, generated outputs remained largely aligned with the training data, exhibiting recognizable stylistic and compositional features inherited from the corpus. Although minor variations and distortions were observed, the DCGAN primarily operated within the formal boundaries of learned visual regularities.\\

Building on this baseline, a multimodal Creative Generative Adversarial Network (CGAN) was subsequently implemented, in which visual generation was conditioned on textual representations drawn from contemporaneous literary sources. This conditioning does not correspond to a simple class-based or attribute-based constraint, but rather to a form of cross-modal conditioning within a shared latent space, enabling structural alignment between textual and visual modalities. As a result, the CGAN exhibited a higher degree of formal variability throughout training. While some early iterations produced outputs that remained partially aligned with the visual corpus, the multimodal conditioning introduced structural tensions that limited direct replication and promoted exploratory recombination.\\

Over continued training, these dynamics facilitated a progressive departure from corpus-bound forms, leading to the emergence of novel visual structures that are not directly traceable to the original data distribution. The objective of this implementation was not to optimize or evaluate creativity, but to observe how creative behavior may emerge within a domain-limited generative system when individual-level and historical-level constraints are jointly enforced.

\subsection{Model Architecture}

The primary model used in this work is a Creative Generative Adversarial Network (CGAN) extended to a multimodal setting. In this configuration, the generator produces visual outputs conditioned on textual representations, while the discriminator evaluates generated images with respect to both visual realism and semantic consistency with the associated textual input.
This multimodal configuration enables the model to learn correlations between visual representations and linguistic descriptions within a shared latent structure. Rather than relying on explicit symbolic rules or handcrafted semantic constraints, the model internalizes cross-modal relationships through adversarial training, allowing textual and visual modalities to become structurally aligned within the generative process.

\subsection{Domain-Limited Corpus Construction}

To instantiate a historically bounded creative environment, both the visual and textual corpora were restricted to the European 18th century.
The visual corpus consists of heterogeneous paintings from the period, including landscapes, still lifes, portraits, and genre scenes. Each image is annotated with descriptive textual metadata, covering elements such as visual composition, depicted objects, stylistic features, chromatic properties, and, when applicable, inferred artistic intentions.

The textual corpus comprises literary and critical texts produced during the same century. These include excerpts from novels, essays, and critical writings contemporaneous with the visual material. The use of period-consistent texts ensures that the linguistic modality remains aligned with the same historical and cultural context as the visual corpus.
Together, these corpora define a domain-limited dataset corresponding to a single historical period, implementing the Zeitgeist component $\textit{Z}_S$ introduced in Section 3.

\subsection{Conditioning and Prompting Strategy}

As the CGAN requires conditioning inputs to drive generation, textual prompts were selected exclusively from 18th-century literary sources contained within the corpus. This choice was motivated by the intention to minimize external subjectivity and authorial intervention.
Rather than introducing contemporary descriptions or manually designed prompts, sentences drawn directly from the period’s literature were used as conditioning signals. This strategy ensures that both conditioning and generation operate within the same bounded informational domain, preserving internal coherence between modalities and historical context.

\subsection{Training Protocol}

The model was trained over approximately 200 iterations, with the number of epochs per iteration varying between 50 and 200, and averaging approximately 90–100 epochs. Training parameters were adjusted across iterations to observe changes in generative behavior over time.
During early iterations, generated outputs remained visually and stylistically close to the training corpus, exhibiting distortions attributable to adversarial training dynamics and dataset heterogeneity. As training progressed, the model increasingly departed from direct replication of corpus elements, producing outputs that exhibited novel compositional structures and stylistic deviations.
These shifts were not explicitly enforced through additional constraints or objectives, but emerged through continued adversarial training and parameter evolution.

\subsection{Observational Categories}

Generated outputs were analyzed qualitatively and grouped into two descriptive categories:

\begin{itemize}
	\item \textbf{Close-to-corpus samples}, which preserve recognizable stylistic and compositional features present in the training data.
	\item \textbf{Emergent samples}, which diverge significantly from corpus examples while remaining internally coherent within the learned domain.
\end{itemize}

These categories are observational and descriptive rather than evaluative. No creativity scores or quantitative novelty measures were applied.

\subsection{Visual results}

Figures 1-2-3 present representative outputs generated by the DCGAN baseline across different epochs within a single training iteration. All samples were produced using the same model configuration and visual corpus, with the only variable being the training epoch. Early-epoch outputs exhibit low-frequency structure formation and blurred reconstructions of corpus-level motifs, reflecting the initial convergence of the generator toward dominant visual features present in the dataset.\\

As training progresses, later-epoch outputs show improved local coherence and sharper textural organization. However, the overall generative behavior remains highly conservative: compositional layouts, chromatic distributions, and figurative structures remain strongly aligned with the original 18th-century visual corpus. While minor distortions and recombinations occur, the DCGAN predominantly interpolates within the learned visual manifold and does not exhibit significant structural deviation from inherited patterns.

\begin{figure*}[htp]
    \centering
    \includegraphics[width=10cm]{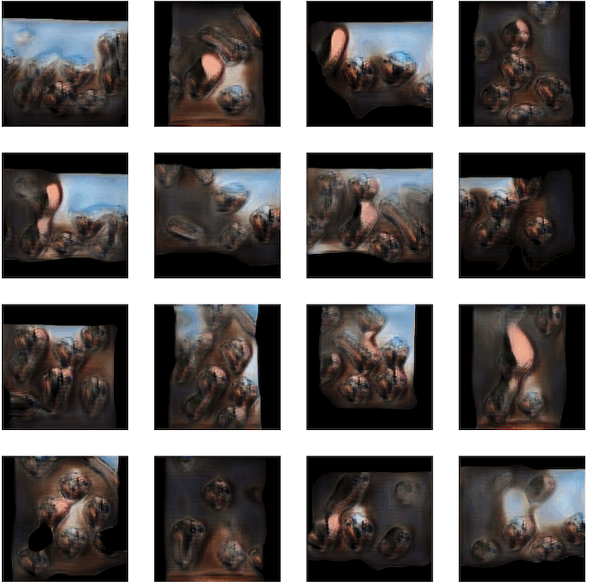}
    \caption{DCGAN (early epoch) - Representative outputs generated by the DCGAN during early training epochs. Generated images exhibit blurred reconstructions and low-frequency combinations of dominant corpus features, reflecting initial convergence toward dataset-level visual statistics.}
    \label{fig:dcgan}
\end{figure*}

\begin{figure*}[htp]
    \centering
    \includegraphics[width=10cm]{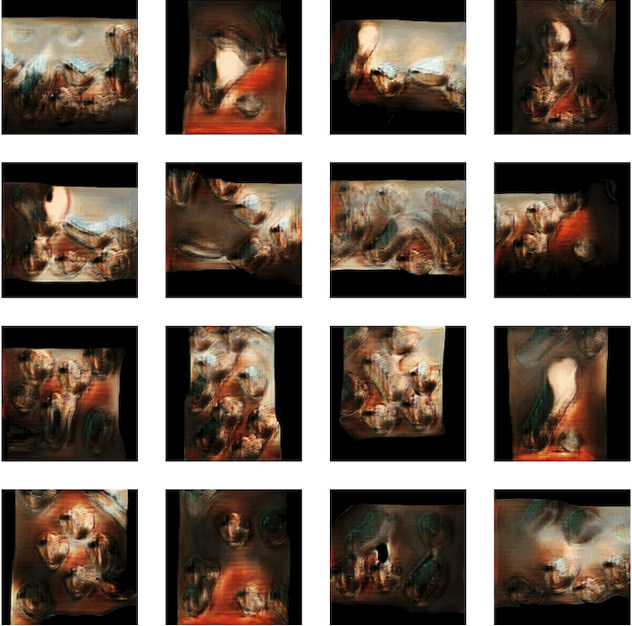}
    \caption{DCGAN (late epoch) - Representative outputs generated by the DCGAN at later training epochs within the same iteration. While visual coherence and textural sharpness improve, generated samples remain strongly aligned with the stylistic and compositional structures of the 18th-century corpus.}
    \label{fig:galaxy}
\end{figure*}

\begin{figure*}[htp]
    \centering
    \includegraphics[width=10cm]{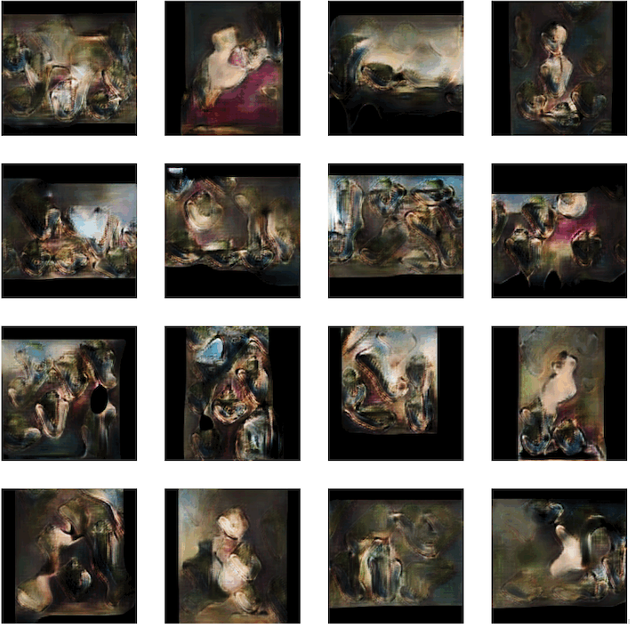}
    \caption{DCGAN (final epoch, same iteration - DCGAN outputs at the final epochs of a single training iteration. Despite increased local consistency, the generative behavior remains conservative, with outputs largely confined to interpolations within the learned visual manifold.)} 
    \label{fig:dcgan_final}
\end{figure*}

In contrast, Figure 4 presents a batch of samples generated by the multimodal CGAN at different stages of training. These samples illustrate a marked departure from corpus-aligned visual structures. The introduction of cross-modal text–image conditioning induces latent-space tensions that prevent direct visual replication and promote the emergence of novel configurations. Generated outputs display abstract shapes, non-referential color fields, and formal organizations that cannot be directly mapped to corpus instances.
This comparison highlights a structural difference between unimodal and multimodal generative dynamics. While the DCGAN stabilizes around corpus-level regularities, the CGAN explores a broader region of the generative space, producing outputs that reflect neither direct imitation nor unstructured noise, but rather emergent patterns arising from cross-modal interaction and adversarial optimization.

\begin{figure*}[htp]
    \centering
    \includegraphics[width=15cm]{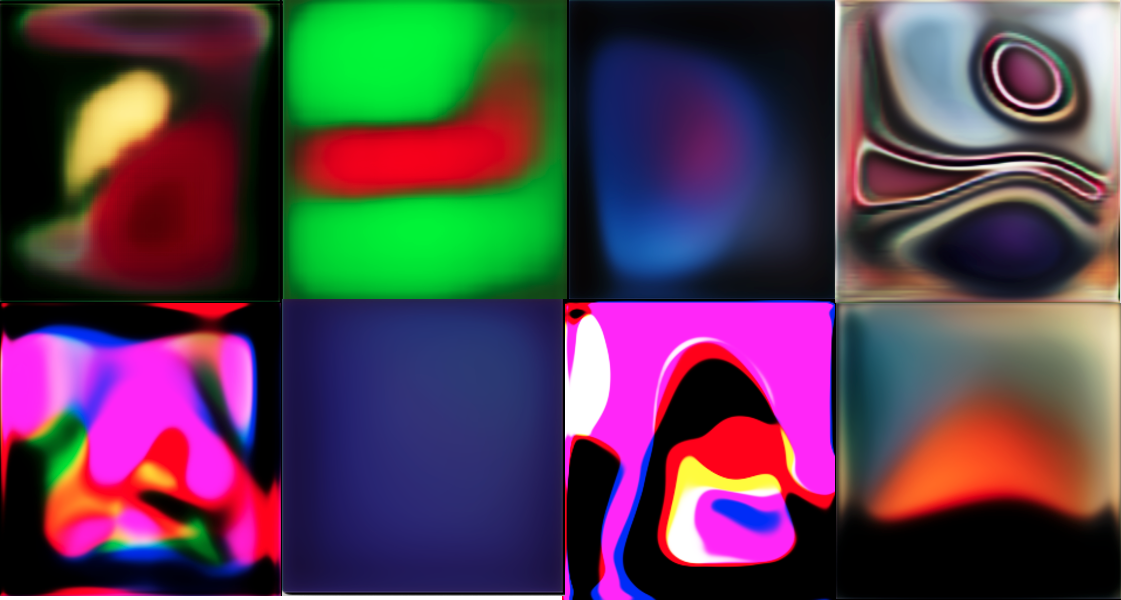}
    \caption{CGAN (emergent samples) - Batch of samples generated by the multimodal CGAN across different training iterations. Outputs exhibit significant formal divergence from the original corpus, including abstract configurations and novel structural patterns, illustrating the impact of cross-modal text–image conditioning on generative dynamics.}
    \label{fig:batch_sample}
\end{figure*}
\section{Discussion and Perspectives}

The experimental observations reported in Section 4 suggest that creativity, when framed as an emergent property of domain-limited generative systems, may generalize beyond the specific architecture and modality explored in this work. Rather than being tied to a particular model class, the proposed framework defines a set of structural conditions that can, in principle, be instantiated across a wide range of algorithms and environments.

\subsection{Extension to Larger and Alternative Generative Systems}

Although this study focuses on a multimodal CGAN trained on textual and visual data, the underlying formal structure is not architecture-specific. The decomposition into historical context, individual world modeling, pattern accumulation, and residual arbitrarity can be implemented in other multimodal generative systems, including transformer-based architectures, diffusion models, or hybrid frameworks.
Moreover, the framework may be compatible with evolutionary and genetic algorithms, where pattern accumulation, environmental constraints, and stochastic variation are already central mechanisms. In such settings, the proposed formulation provides a conceptual lens for interpreting how creative solutions may emerge from iterative variation and selection within bounded domains.

\subsection{Creativity and Generalization Beyond Known Domains}

One of the persistent challenges in computational creativity and artificial intelligence more broadly is the limited capacity of systems to address problems that fall outside their training distributions. The present framework suggests that this limitation may be partially addressed by shifting the focus from task-specific optimization to the construction of generative spaces that support exploratory behavior.
By explicitly separating structured constraints from irreducible arbitrarity, the model allows for deviations that are not fully determined by prior data. Such deviations may enable systems to explore solution spaces that are not explicitly represented in their training sets, thereby supporting forms of problem-solving that extend beyond direct interpolation.

\subsection{Toward Embodied and Robotic Systems}

A natural extension of this work lies in embodied systems and robotics. While the present model operates within a purely representational domain, physical interaction with an environment introduces empirical feedback that cannot be fully captured through static datasets.
Embedding the proposed framework in robotic systems would allow creativity to emerge not only from symbolic or perceptual data, but from direct interaction with the world. Such interaction could ground pattern accumulation and worldview formation in sensorimotor experience, potentially supporting more adaptive and autonomous forms of creative behavior.

Multimodal generative models provide a form of structural coherence by inducing shared latent spaces in which heterogeneous sensory modalities are jointly represented. In such spaces, visual, textual, or auditory signals are mapped onto a common representational substrate, enabling cross-modal associations and internal consistency. This representational capacity is analogous to the structural component of a human worldview, in which diverse perceptual inputs are organized within a coherent internal model.
However, multimodality alone does not provide causal grounding. In the absence of embodiment, a generative system lacks access to sensorimotor contingencies and cannot establish systematic relationships between perception, action, and consequence. Without such feedback, representations remain detached from empirical interaction and cannot be updated through experiential correction.
Embodied systems, such as robotic agents, restore this missing loop by coupling perception with action in a physical environment. Through interaction, the system can test internal representations against external constraints and incorporate the outcomes of its actions into subsequent generative and inferential processes. This enables the transition from purely representational coherence to experience-grounded adaptation.

In this context, self-deductive or inference-based mechanisms play a complementary role by integrating new experiences into an evolving internal structure. Deductive processes allow the system to reconcile prior representations with novel observations, supporting the formation of a coherent and dynamically updated worldview.

Taken together, multimodality, embodiment, and deduction can be understood as three complementary layers of a creative and cognitive system: multimodality provides the representational layer, embodiment supplies the experiential layer, and deduction constitutes the interpretative layer. Their interaction defines the structural conditions under which a machine could develop a functional form of Weltanschauung, grounded in both internal coherence and external experience.

\subsection{Constraints, Autonomy, and Creative Capacity}

An important implication of this work concerns the role of constraints in creative systems. While domain limitation is essential for coherence, excessive constraints at the generative level may suppress exploratory behavior. Both in human development and in algorithmic systems, over-constraining learning processes can lead to rigid patterns and reduced adaptability.

The framework presented here suggests that creativity benefits from a balance between structured constraints and freedom of exploration. Preserving space for arbitrarity and deviation may be essential for enabling systems to escape narrow solution regimes and develop autonomous problem-solving capacities.

\section{Conclusion}

This work proposed a non-anthropocentric framework for creativity in artificial systems, grounded in the interaction between structured constraints, individual internal organization, and irreducible contingency. Rather than treating creativity as an evaluative property of outputs, the paper formalized it as an emergent phenomenon arising within domain-limited generative systems. A mathematical formulation was introduced to describe creativity as the interaction between historical context, individual world modeling, accumulated patterns, and a residual arbitrarity term. This framework was then instantiated through a multimodal CGAN trained on an 18th-century textual and visual corpus, allowing creative behavior to be observed as a generative process rather than measured post hoc.

Within this perspective, creativity is defined independently of consciousness, emotion, or biological intention. It is understood as the capacity of a system—biological or artificial—to produce novel structures from within its own internal organization, constraints, and modes of interaction with its environment. The concepts of alterity and idiosyncrasy capture this idea: creative behavior emerges from the fact that every system processes reality through a specific architecture, representational structure, and experiential history. Novelty, in this sense, is not an external criterion but a consequence of structural difference.
The experimental observations suggest that when a generative system is immersed in a coherent cultural domain and allowed to iterate over time, it may transition from pattern reproduction to the emergence of novel and internally coherent structures. Such behavior supports the view that creativity functions as a mechanism of adaptation and exploration, rather than as a domain-specific or exclusively artistic capability.

From a broader perspective, these results have implications for the development of more autonomous artificial systems. Intelligence cannot be reduced to the execution of predefined tasks or the optimization of fixed objectives. The capacity to generate solutions in ill-defined, novel, or underdetermined situations requires mechanisms that support exploratory generation, pattern reinterpretation, and structural novelty. In this sense, creativity is not a byproduct of intelligence, but one of its enabling mechanisms.

While this work does not claim to achieve artificial general intelligence, it suggests that formalizing creativity as an emergent generative process may be a necessary step toward systems capable of adaptive generalization. By framing creativity as a structural and computational phenomenon, rather than an anthropocentric attribute, this approach opens new directions for the design of generative models, multimodal systems, and embodied agents capable of expanding their internal representations and problem spaces over time.

\nocite{*}
\printbibliography 


\end{document}